\documentclass[10pt]{article} 
\usepackage[accepted]{tmlr}

\usepackage{amsmath,amsfonts,bm}

\def\eqref#1{equation~\ref{#1}}

\def\1{\bm{1}}

\DeclareMathAlphabet{\mathsfit}{\encodingdefault}{\sfdefault}{m}{sl}
\SetMathAlphabet{\mathsfit}{bold}{\encodingdefault}{\sfdefault}{bx}{n}

\usepackage{graphicx}
\usepackage{hyperref}
\usepackage{url}
\usepackage{multirow}

\title{Identifying and Mitigating Model Failures through Few-shot CLIP-aided Diffusion Generation}

\author{\name Atoosa Chegini 
\email atoocheg@umd.edu \\
      \addr Department of Computer Science\\
      University of Maryland, College Park
      \AND
      \name Soheil Feizi \email sfeizi@umd.edu \\
      \addr Department of Computer Science\\
      University of Maryland, College Park}

\begin{document}

\maketitle

\begin{abstract}
Deep learning models can encounter unexpected failures, especially when dealing with challenging sub-populations. One common reason for these failures is the occurrence of objects in backgrounds that are rarely seen during training. To gain a better understanding of these failure modes, human-interpretable descriptions are crucial for further analysis and improvement which is expensive. In this study, we propose an end-to-end framework that utilizes the capabilities of large language models (ChatGPT) and vision-language deep models (CLIP) to generate text descriptions of failure modes associated with spurious correlations (e.g. rarely seen backgrounds) without human-in-the-loop intervention. These descriptions can be used to generate synthetic data using generative models, such as diffusion models. The model can now use this generated data to learn from its weaknesses and enhance its performance on backgrounds that are uncommon for each class of data. Our approach serves as a broad solution, promising progress in comprehending model failure modes and strengthening deep learning models across a wide range of failure scenarios (e.g. bacckgrounds, colors) automatically in a few-shot manner. Our experiments have shown remarkable \textbf{improvements in accuracy ($\sim \textbf{21\%}$)} on hard sub-populations (particularly for wrong background association) across $40$ different models, such as ResNets, EfficientNets, DenseNets, Vision Transformer (ViT), SwAVs, MoCos, DINOs, and CLIPs on various datasets such as ImageNet-1000, CIFAR-10, and CIFAR-100.
\end{abstract}

\begin{figure*}[h]
        \centering
        \includegraphics[trim=0cm 0cm 0cm 0cm, clip, width=\linewidth]{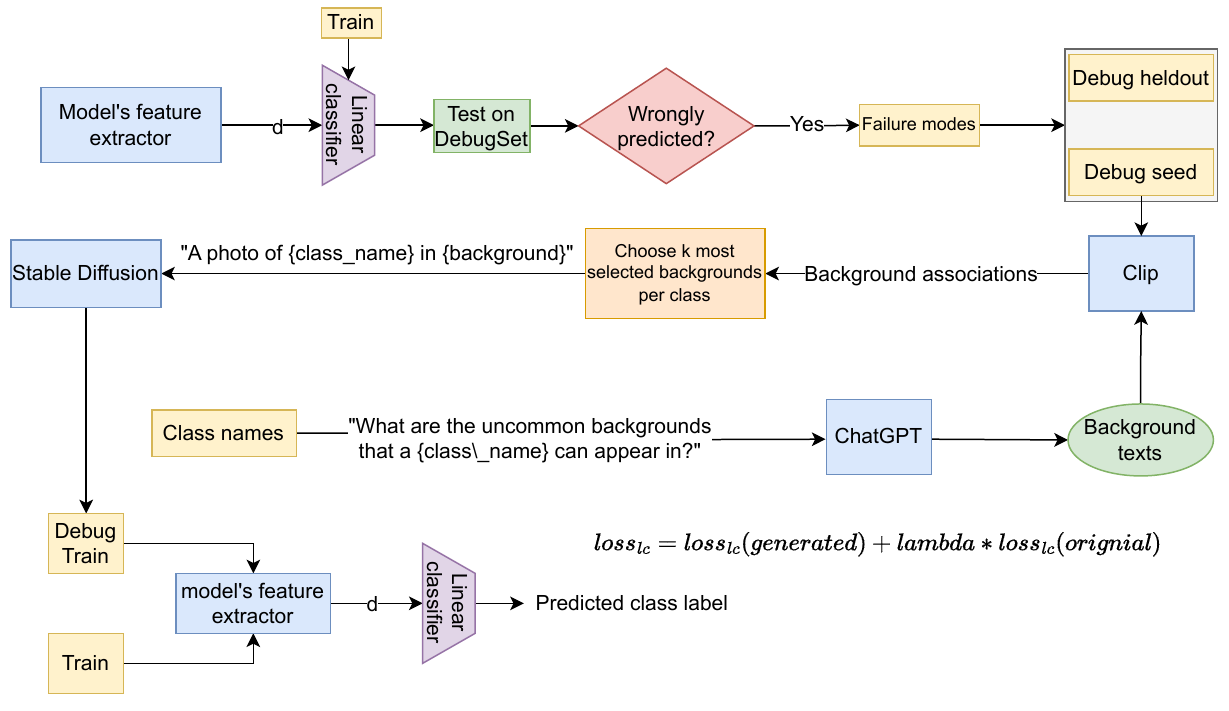}
        \label{fig:framework}
    \caption{A summary of our approach: For a model based on the wrongly predicted debug samples (failure samples), we form two sets - debug seed and debug heldout. We use the debug seed set to address the model's failures by inputting them to CLIP \cite{radford2021learning}, along with a set of backgrounds obtained from ChatGPT where objects are less likely to occur. We then obtain a set of backgrounds and remove redundancies, and generate synthetic data by inputting the prompt "A photo of \{class\_name\} \{background\}" to Stable Diffusion \cite{rombach2021high}. With this synthetic data that precisely captures the model's failure modes, we can now debug the model's predictions on other test data by training a very low-cost network on top of our model, which assigns different weights to the original data and the generated data.}
    \label{fig:main_fig}
\end{figure*}

\section{Introduction}

The quality of training data directly impacts the performance and robustness of machine learning models. Despite careful curation of training data, models can still exhibit failure modes where their performance deteriorates in specific sub-populations of data, leading to misclassifications or inaccurate predictions \cite{jiang2018mentornet, arpit2017closer}. The failure modes of deep networks can arise from various factors, such as noisy labels \cite{sukhbaatar2014training, jiang2018mentornet, reed2015training}, multi-labels \cite{zhang2018multi}, and spurious correlations \cite{zhou2020towards}, particularly when it comes to distinguishing between objects and their backgrounds \cite{kattakinda2021focus, xiao2020noise}. (See the figure \ref{appendix:failure_examples} in appendix for examples of these failures.)

Similar to how humans use image backgrounds as cues for object recognition, studies have shown that machine learning models also rely on backgrounds when making decisions. In some cases, models may prioritize backgrounds to the point of overlooking important object features for classification \cite{zhang2007local, ribeiro2016should, sagawa2019distributionally}.

Various approaches have been attempted to address failure modes caused by spurious background associations, but many of them are inadequate in addressing all aspects of the problem. Some approaches involve human-in-the-loop interventions \cite{mitchell2021fast, santurkar2021editing}, which is labor-intensive and challenging to apply to large-scale operations. Moreover, many of them only target a limited set of failure modes, neglecting the comprehensive spectrum of potential failures \cite{barbu2019objectnet, hendrycks2021many, hendrycks2019benchmarking, hendrycks2021natural, kattakinda2021focus}. Additionally, some of the existing work lacks clear descriptions of model failures in a human-understandable manner, posing challenges in terms of interpretability and validation.

Alongside research on identifying failure modes, there are several debugging approaches that seek to utilize these failure modes to enhance the accuracy of machine learning models. These include creating supplementary datasets with failure samples to help the model learn robust features \cite{xiao2020noise, singla2022data}, or modifying the model's parameters to incorporate information from identified failure modes \cite{rame2022diverse}. These studies lack failure mode descriptions that are easily understandable for humans, making it challenging to interpret their validity.

\section{Our contribution}

This research leverages recent generative models, large language models, and CLIP to introduce an automated framework addressing failure modes (spurious correlations) in diverse task-specific deep learning models. The framework, outlined in Figure \ref{fig:main_fig}, answers key questions such as identifying and rectifying spurious associations leading to model failure, utilizing these failure modes to enhance model performance, exploring patterns in failure modes across a model group, and using a single set of auxiliary data to simultaneously improve a subgroup of models.

To summarize, our approach initially identifies all model failures on a specific subset, denoted as \textbf{\textit{DebugSet}}, within the validation set. We then pinpoint spurious correlations, such as background associations, for each dataset class by querying ChatGPT with "What are the uncommon backgrounds that a {class\_name} can appear in?" and remove redundacies after obtaining all uncommon backgrounds. Subsequently, a zero-shot classification using CLIP identifies the background for each failure among all the uncommon backgrounds. To enhance model performance, we generate $k$ artificial images with prompts like "[class\_name] in [background\_name]" and incorporate this supplementary data into the original training set. In the second phase, we demonstrate that models with similar architectures exhibit analogous failures, allowing for efficient troubleshooting of a model group using a single set of generated auxiliary data. This approach proves both time and memory efficient. The results of our experiments, detailed in section \ref{section:Experiments}, underscore the effectiveness of this straightforward method in achieving interpretability and debugging goals.

Our paper presents several contributions to the field of machine learning model failure analysis and debugging. These contributions include, but are not limited to, the following:

\begin{itemize}

    \item \textbf{Generalizability}: Introducing a comprehensive end-to-end framework that interprets and rectifies failures arising from specific spurious associations, such as incorrect background, color, and size correlations, which can contribute to any model inaccuracies.
    
    \item \textbf{Failure Inspection}: Identification of spurious background association failure modes of $\sim 40$ multiple models on ImageNet in an interpretable manner (section \ref{section:failure_individual}), and exploring common patterns in failure modes among individual models with same architectures (section \ref{section:failure_all}) automatically and without any human intervention.

    \item \textbf{Failure Mitigation}: Improving the performance of individual models on challenging sub-populations (\ref{section:debugging_individual}), and boosting the performance of model subsets by employing a unified set of auxiliary data, leveraging shared failures to enhance efficiency in both time and memory usage (section \ref{section:debugging_all}).

    \item \textbf{Collective Failure Mitigation}: Enhancing subsets of models' performance through a unified set of auxiliary data owing to their shared failures which saves time and memory. To the best of our knowledge, this work represents the first effort to collectively address failures within a subgroup of models simultaneously. (section \ref{section:debugging_all}).

    \item \textbf{Dataset creation for debugging ImageNet, Cifar10, and Cifar100}: We created a dataset that will improve models' performance on background and color associated failure modes on these three datasets.
\end{itemize}

\section{Related work}
\subsection{Failure mode detection}

Numerous studies have been conducted to detect machine learning model failure modes. As previously mentioned, some involve human-in-the-loop methods, where failure examples are reviewed to identify common patterns \cite{mitchell2021fast, santurkar2021editing}. Others adopt automated approaches by introducing frameworks that effectively capture model failures \cite{chung2019slice, singla2021understanding, nushi2018towards, singla2021salient, wong2021leveraging, wu2019errudite, zhang2018manifold, jain2022distilling}. For instance, \cite{chung2019slice} employs a technique that slices the validation data to isolate sections where the model performs poorly. \cite{singla2021understanding} identifies visual attributes that lead to inadequate performance when present or absent. \cite{jain2022distilling} identifies and represents model failures as directions in the latent space, and \cite{eyuboglu2022domino} that proposes an evaluation framework to systematically compare (slice discovery method) SDMs across diverse slice settings by generating captions for hard sub-populations. Distinguishing itself from existing methodologies, our approach provides enhanced generality by \textbf{permitting the explicit selection of the spurious correlation targeted for mitigation}. For instance, although the approach presented by Kattakinda et al. \cite{kattakinda2022invariant} effectively tackles spurious correlations tied to foreground and background features by learning disentangled representations, it encounters difficulties when confronted with a wider spectrum of spurious correlations, e.g. color. This is due to the inherent challenge of learning disentangled representations for many spurious correlations in isolation from the foreground object.

\subsection{mitigation of hard subpopulations and interpretability of models}

Several methodologies leverage extracted failure modes to debug and enhance the performance of deep learning models. Singla et al. \cite{singla2022data} introduce a framework that identifies visually similar images to model failures and incorporates them as new data for debugging various models. Kattakinda et al. \cite{kattakinda2022invariant} focus on learning invariant features for foreground and background by penalizing the mutual information between the features and background/foreground labels. This approach contributes to robust model training, particularly by addressing issues related to spurious correlations.

In the context of data generation, Bansal and Grover \cite{bansal2023leaving} and Wiles et al. \cite{wiles2022discovering} use generated data to diversify training datasets. However, it's essential to note that their methods do not specifically target failure modes like spurious correlations. They rely on class names and general captions for generating auxiliary data, which may not be tailored to address specific failure modes.

Moreover, Wiles et al. \cite{wiles2022discovering} propose a bug discovery approach using off-the-shelf image generation and captioning, contributing to the interpretability of model predictions. On the other hand, Jain et al. \cite{jain2022distilling} leverage Support Vector Machines (SVMs) to distill model failures as directions in latent space, focusing on latent representations of model failures.

In comparison to existing methodologies that address failure modes on specific datasets, our framework introduces two noteworthy contributions. Firstly, \textbf{it achieves enhanced model performance with significantly fewer generated examples, (5 for each failure)}. Secondly, \textbf{our experiments extend to collective debugging, demonstrating the ability to improve a subset of model failures by generating a single auxiliary artificial dataset based on only one model's failures}. This is particularly valuable given our observation that models within the same categories exhibit similar failures, a phenomenon also noted in \cite{wiles2022discovering}.

Moreover, our approach stands out for its efficiency, eliminating the necessity for complete model retraining or fine-tuning. We exclusively focus on retraining the linear head for classification, streamlining the failure mode mitigation process.

\section{Main method}

\subsection{Failure-mode detection}
A common reason for accuracy drops during inference is the model's learned spurious correlations from training. For example, Associating objects with backgrounds, a spurious correlation, can hinder the model's ability to learn objects themselves. This challenge arises when the model encounters objects in unfamiliar backgrounds during testing, notably in computer vision tasks where backgrounds define object context. To tackle this, introducing the model to a range of scenarios that address the particular failure mode (such as color or background associations) we aim to mitigate, can improve its ability to identify objects in different contexts, and avoid correlating the objects and their changable features (e.g. color) or contexts (e.g. background).

In this work we explain how we use this framework for wrong background associations, however it can easily be applied to all other spurious correlations that models may mistakenly learn.

To address and rectify failures attributed to backgrounds, we utilize the feature extractor for each model on the datasets, generating a feature vector for each data point. The subsequent linear head atop this feature extractor is responsible for executing the classification task. Instances where the model makes incorrect predictions form a set termed the \textbf{\textit{"debug set"}}. This debug set serves as a tool for identifying and resolving failure modes, as it comprises all examples where the model fails. While these failures may stem from various factors, our experiments underscore the significance of mitigating incorrect background associations, as it significantly improves the performance of all models.

\begin{table}[h]
    \centering
\begin{tabular}{||c|c||} 
 \hline
 Class name & Uncommon backgrounds\\
 \hline \hline
 
Sea lion & Desert, Rain forests, Urban Areas, Polar Ice Caps, \\ & Caves, Grasslands, Volcanic Areas\\ \hline
Siberian husky & Jungle Canopies, In sky, Caves, Underwater, \\
& Indoor Spaces, Marshlands, Tropical Rainforests\\ \hline
croquet ball & Mountain Peaks, Busy Streets, Frozen Lakes, \\
& Underneath Building Foundations, Subway Tunnels, in a restaurant\\ \hline
lipstick, lip rouge & Gyms and Fitness Centers, Swimming Pools, \\
& Medical Facilities, Construction Sites, Sports Events, Military Training \\ \hline
 \hline
\end{tabular}
 \caption{Examples of suggested uncommon backgrounds for a class of data by ChatGPT}
\label{table:chatgpt}
\end{table}

\subsection{Failure-mode textualization}
Vision-language models are popular as they can provide more comprehensive understanding of complex phenomena by combining information from different modalities like text, images, and audio, enabling them to interpret data in a more human-readable form \cite{lu2019vilbert, chen2018three, mithun2020neural}.

Understanding failure modes is critical for validating proposed debugging methods. By identifying the causes of failure, we can improve our models and refine our data collection methods. For each class\_name in our dataset, we first prompt ChatGPT "What are the uncommon backgrounds that a {class\_name} can appear in?" and filter out the redundant suggested backgrounds. Some examples can be seen in Table \ref{table:chatgpt}. Then, we use CLIP \cite{radford2021learning} to interpret failure modes by splitting the failures from the \textbf{\textit{debug set}} into two sets called \textbf{\textit{debug seed}} and \textbf{\textit{debug heldout}}. We then perform zero-shot classification by passing \textbf{\textit{debug seed}} along the set of class\_wise uncommon backgrounds proposed by ChatGPT to a CLIP model, so for each data point, CLIP will opt for the background that is more likely to be the actual background of the object shown in the image. For each data class, we then pinpoint the $\textbf{\textit{k}}$ most frequently selected backgrounds by CLIP, which the model failed to classify. This will provide valuable insights into the ways how a model may fail when confronted with a particular selected spurious association.

\begin{figure}
        \centering
        \includegraphics[trim=0.4cm 0cm 0cm 0cm, clip, scale=1.1]{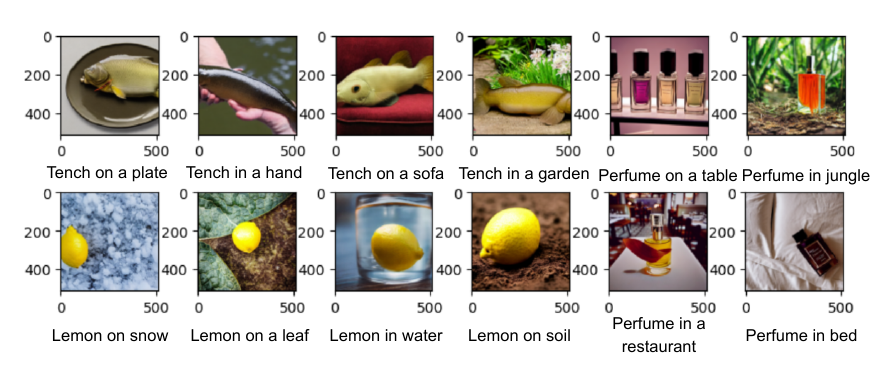}
    \caption{Examples of generated data by Stable Diffusion} 
    \label{fig:synthetic_data_examples}
\end{figure}

\subsection{Generating synthetic data}
By leveraging the detected backgrounds of failures by Clip, we can both interpret those failures, and use them to debug models. For instance, in the case of the Imagenet class "tench," errors predominantly occur when the fish is held by a person's hand, a scenario rarely encountered during training. To address this, a generative model like Stable Diffusion \cite{ho2020denoising} can be used to create images that familiarize the model with diverse object contexts. For the "tench" class, we generate data by inputting the prompt "tench in a hand" to the Stable Diffusion. Examples of such generated data are presented in figure \ref{fig:synthetic_data_examples}

\subsection{Retraining the linear head}
After collecting the additional synthetic data for the failed scenarios, which we call \textbf{\textit{debug\_train}}, we can use it along with our original trainset to debug our models. To achieve this, we only need to train a linear head on top of the model's feature extractor for the classification purpose and not the whole model. It is important to note that we assign different weights to the datapoints from the original\_train set and the Debug\_train set in our linear head training loss \ref{equ:lambda}. This parameter is called \textbf{\textit{lambda}}, and in our experiments, we report its effect on the overall performance of the model. By incorporating the additional debug\_train data and carefully tuning the lambda parameter, we can potentially improve the performance of our models.

\begin{equation}
    L_{cl} = L_{cl}(Original\_train) + \lambda * L_{cl}(Debug\_train)
\label{equ:lambda}
\end{equation}

\begin{figure*}
        \centering
        \includegraphics[trim=0cm 0cm 0cm 0cm, clip, width=\linewidth]{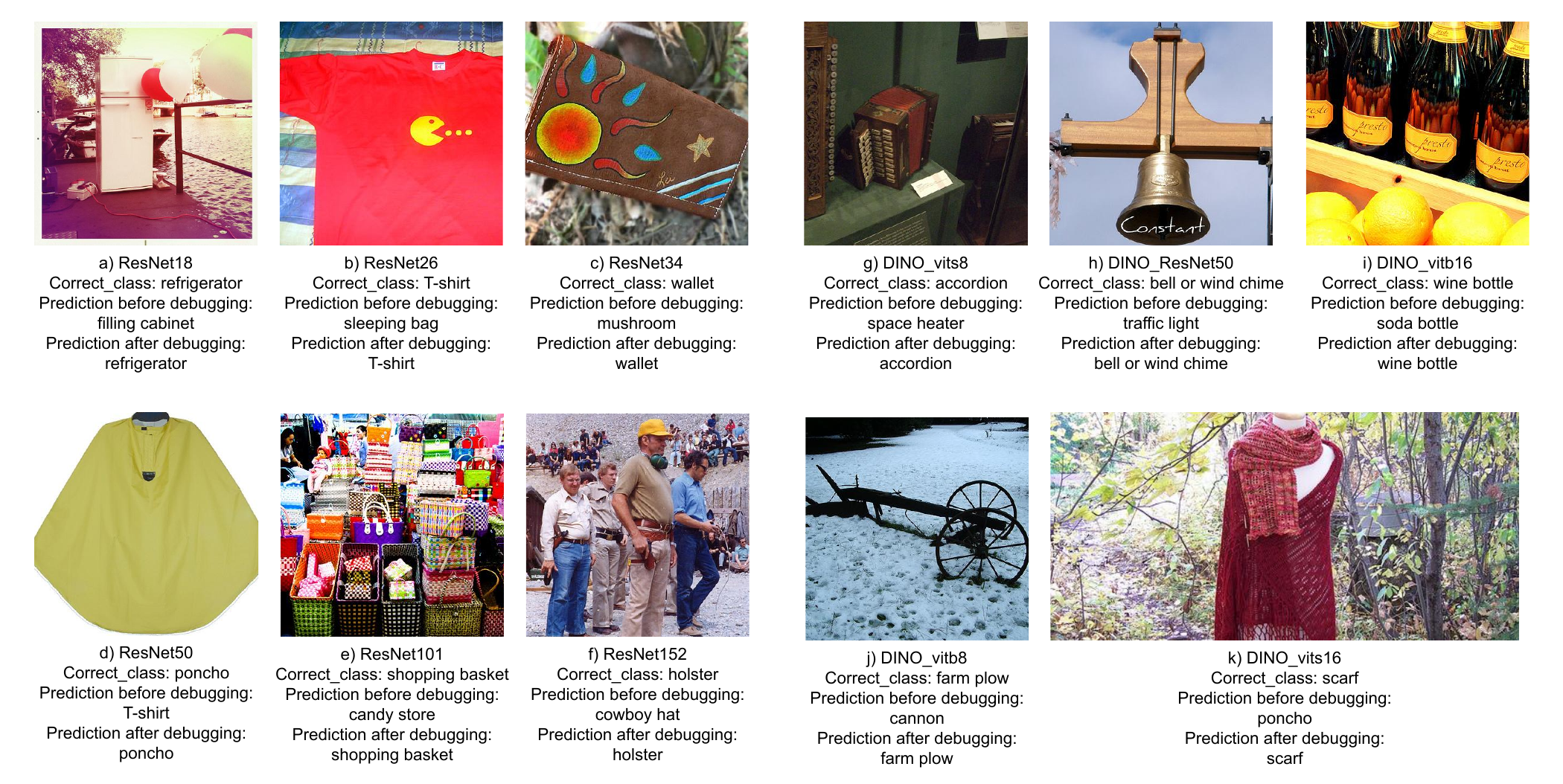}
    \caption{Some examples of failure modes of ResNets and DINOs 
    \label{fig:failures_resnet_dino}}
\end{figure*}

\section{Experiments}
\label{section:Experiments}
\subsection{Setting}
We load our datasets and use their training data to train the linear head on top of the models' extracted features. For ImageNet, we use $30$ data points per class and the overall $30000$ training images. We choose $30000$ images out of $50000$ of Imagenet's validation set as our \textit{debug\_set}, and the remaining $20000$ samples will be considered for the testing process. Each image's resolution is $224 * 224$, and the task performed here is the classification task on Imagenet classes.
Other settings include hyper-parameters that we used can be seen in table \ref{table:shared_params} in the appendix.

We tested our method on $40$ different models including ResNets \cite{he2016deep}, EfficientNets \cite{tan2019efficientnet}, DenseNets \cite{huang2017densely}, Vision Transformer (ViT)  \cite{dosovitskiy2021image}, SwAVs \cite{caron2020unsupervised}, MoCos \cite{he2019momentum}, DINOs \cite{caron2021emerging}, and Clips \cite{radford2021learning}. The exact list of used models can be seen in table \ref{table:models_list} in the appendix. For the sake of space, we only show experiments on DINO and ResNet models, and experiments for other models can be found in appendix \ref{appendix:B}.

We split the detected failures of \textit{debug set} in half. The first half, \textit{debug seed} will be used for debugging. For obtaining uncommon backgrounds, we use ChatGPT $3.5$. The CLIP model we use for choosing backgrounds for data points is ViT-B\/32 CLIP.
For generating synthetic data, we use Stable Diffusion $V1-5$ imported from the diffusers package.

\subsection{Failure inspection}
The initial stage of our framework involves analyzing how various models fail to classify objects on different datasets. To accomplish this, we use the CLIP model to identify backgrounds on which models struggle to classify objects. This results in captions that describe failures related to rare backgrounds. In the following stage, we examine these identified failures and explore how the generated captions help us to recover from them.
We investigate results for both \textbf{individual and Collective failure inspection}.

\begin{figure}
        \centering
        \includegraphics[trim=0.26cm 0cm 0cm 0cm, clip, scale=0.6]{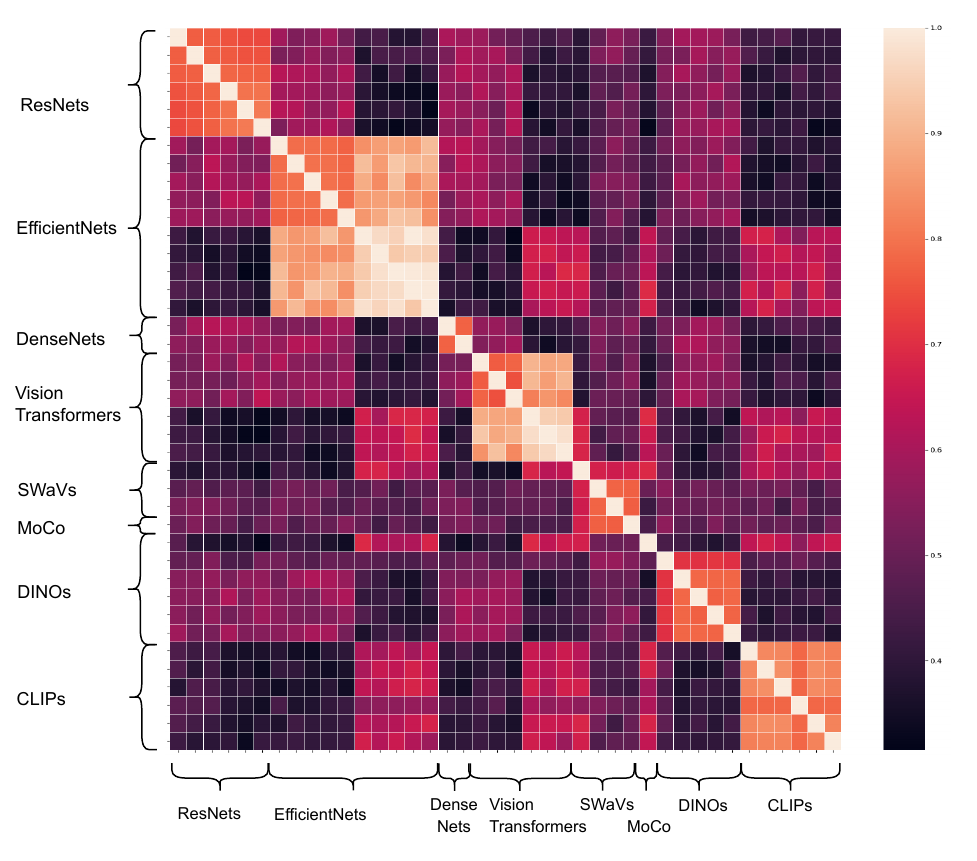}
    \caption{Comparing failures of all models \textbf{(Intersection/Union)}.We observe that models belonging to the same categories tend to exhibit more comparable failures.}
    \label{fig:failures_for_all_models}
\end{figure}

\subsubsection{Individual Failure Inspection}
\label{section:failure_individual}
In figure \ref{fig:failures_resnet_dino}, we show some instances where ResNet and DINO models have failed and see that these failures are due to wrong background association. In this figure, the six images on the left (a-f) are examples of Resnets' failures, and the five images on the right (g-k) are failure modes' of DINO models. For example, image \textbf{\textit{c}} shows \textit{"a wallet in jungle"}, which can be regarded as an uncommon background for this object. As a result, the ResNet34 model is unable to classify it accurately and instead predicts a \textit{"mushroom"} which is more likely to be found \textit{"in garden"}, especially under a plant, despite having no resemblance to the actual object in the image. Similarly, image \textbf{\textit{h}} illustrates \textit{"a bell or wind chime in sky"}, which is uncommon since \textit{"bell"} is more likely to be seen with other backgrounds such as \textit{"a door, a building or a wall"}. Therefore, the DINO\_reset50 model mispredicts as \textit{"traffic light"} because \textit{"traffic light"} is more common to be seen \textit{"in a sky background"}.

In general, our approach is capable of addressing failure scenarios originating from uncommon backgrounds of objects. Analyzing the relevant backgrounds allows us to readily understand the cause of such failure instances.

\subsubsection{Collective Failure Inspection}
\label{section:failure_all}

Within this section, we will conduct a comparison of the failure modes for all models to assess their alignment. Our aim is to determine the extent to which failures are consistent across models. While numerous studies have focused on analyzing similarities in the learning process and representations of different models, such as \cite{raghu2021vision} that demonstrated the similarity between convolutional neural networks (convnets) and other convnets, as well as the similarity between vision transformers (ViTs) and other ViTs, our focus is on investigating whether models also fail in similar ways. This will enable us to gain a deeper understanding of how to address the issue of failures in a more generalized manner without taking the specific model into consideration.

The failures of different models in various categories are compared in figure \ref{fig:failures_for_all_models} by computing the \textit{intersection over union} of the failures. It can be observed that models within the same category fail in more similar samples. Typically, the failures between models from the same category are over $80\%$ similar (e.g. CLIPs and EfficientNets). Among all $40$ models, the intersection of failure modes is above $40\%$, indicating that models tend to fail in very similar ways, even with different architectures. This will raise the question of "how to utilize this similarity in failures to enhance a group of models' performance?" which we will explore more in section \ref{section:debugging_all}.

It is pertinent to mention that \cite{wiles2022discovering} has also recognized patterns of consistencies in failures among models within the same category. however, we take a step further and delve into leveraging these consistencies to systematically mitigate shared failure modes.

\subsection{Failure Mitigation}

\subsubsection{Individual Failure Mitigation}
\label{section:debugging_individual}

The outcomes from employing our framework are presented in Table \ref{table:individual model acc}. we only included the results for ResNets and DINOs, but we have results for other models (EfficientNets, DenseNets, ViTs, SWaVs, MoCo, and CLIPs) in the appendix.
In Table \ref{table:individual model acc}, we constructed \textit{debug\_seed} and \textit{debug\_heldout} sets to yield zero accuracy for the model, as they are composed of model failures. Post debugging and utilizing \textit{debug\_seed}, we observe substantial improvements in \textit{debug\_heldout} data that we didn't use for debugging, ensuring an unbiased evaluation. This improvement underscores that many failure modes stem from incorrect associations models make between objects and backgrounds. Some might argue that this gain results from the additional data. Thus, we present results for a baseline we term \textbf{Random\_debugging}. This baseline similarly uses \textit{debug\_seed} and \textit{debug\_heldout}, then generates synthetic data using only class names (prompts are structured as \textit{"A photo of [class\_name]"}). This comparison illustrates that the improvement of our method arises from considering background information. In the outcomes of Random\_debugging, the improvement over debug\_seed and debug\_heldout is roughly equivalent since no information from the background association of either set was utilized. Random\_debugging solely employs class names to generate data. However, this improvement is not on par with the gain achieved by incorporating background information when generating new data.

\begin{figure}[!h]
    \centering
    \includegraphics[trim=0cm 0cm 0cm 0cm, clip, scale=0.55]{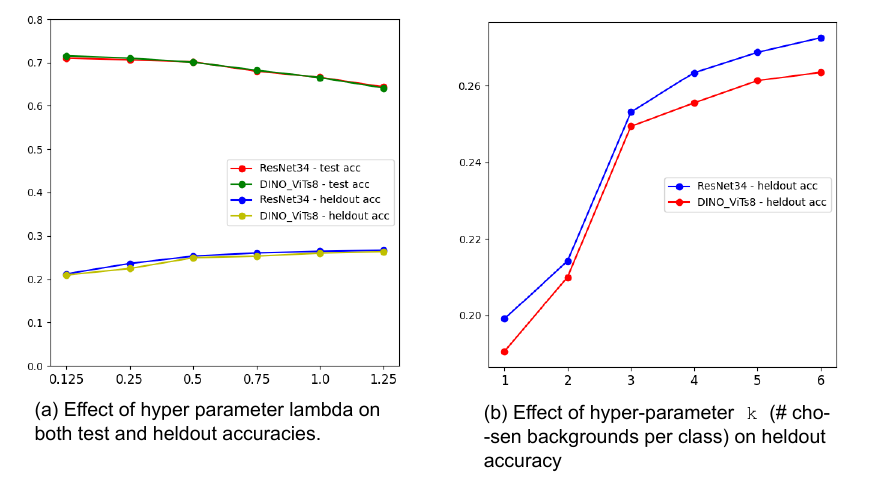}
    \label{fig:ks_lambda}
    \caption{a) As the value of $lambda$ increases, the accuracy on the heldout set improves while the accuracy on the test set decreases. However, there is a specific point \textbf{(0.5)} on the plot where the accuracy of the models on both the test and heldout sets stabilizes. b) Increasing the number of chosen backgrounds per class enhances the accuracy on the heldout set. Considering the high cost of generating additional data, we opt for \textbf{k = 3}, where the plot exhibits a significant slope.}
    \label{fig:hyper_params} 
\end{figure}

It's worth noting that despite incorporating Stable Diffusion-generated data, which could be seen as out-of-distribution samples, a positive impact on model performance remains. This is primarily attributed to the parameter $\lambda$ that controls the contribution of the generated images in our training process. The influence of this parameter is depicted in figure \ref{fig:hyper_params}.

Another crucial hyper-parameter is the number (\#) of generated synthetic data per class. The effect of this hyper-parameter, denoted as \textbf{\textit{k}}, is illustrated in figure \ref{fig:hyper_params}.

The improvement observed in debug\_heldout data surpasses \textbf{$\sim 21\%$} for all models, highlighting the tendency of models to fail in associating backgrounds with objects and utilizing this association to predict objects, neglecting object-specific features. This can be contrasted with the accuracy gain achieved by the \textbf{Random\_debugging} baseline, which is significantly smaller compared to our method.

\begin{table*}
    \centering
\begin{tabular}{||c|c|c|c|c|c|c|c|c||}
\hline
 Models & \multicolumn{8}{ |c|| }{Accuracies}\\ \hline
\hline
 \multirow{3}{*} {Model} & & Accuracy & \multicolumn{3}{ |c|| }{\textbf{Accuracy of}} & \multicolumn{3}{ |c|| }{Accuracy of}\\
    & model & before  & \multicolumn{3}{ |c|| }{\textbf{Individual Debugging}} & \multicolumn{3}{ |c|| }{Random debugging}\\
  category & name & debugging  & \multicolumn{3}{ |c|| }{(ours)} & \multicolumn{3}{ |c|| }{}\\\hline
  
 & & Test & Test & seed & \textbf{heldout} & Test & seed & heldout \\ \hline
\multirow{6}{*}{ResNet} & resnet18 & $0.6236$ & $0.6413$ & $0.2636$ & \textbf{0.2128} & $0.6242$ & $0.1134$ & $0.1129$\\  & resnet26 & $0.6593$ & $0.6671$ & $0.2856$ & \textbf{0.228} & $0.6604$ & $0.09539$ & $0.0904$\\  & resnet34 & $0.7017$  & $0.7165$ & $0.3061$ & \textbf{0.2531} & $0.7099$  & $0.09856$ & $0.08615$\\ & resnet50 & $0.7631$ & $0.7671$ & $0.3444$ & \textbf{0.2717} & $0.7657$ & $0.1102$ & $0.1072$\\ & resnet101 & $0.796$ & $0.8024$ & $0.3574$ & \textbf{0.2656} & $0.7974$ & $0.1132$ & $0.1181$\\ & resnet152 & $0.816$ & $0.8238$ & $0.3609$ & \textbf{0.2817} & $0.8183$ & $0.08207$ & $0.08804$\\ \hline

 \multirow{6}{*}{DINO} & ViTs8 & $0.6977$ & $0.7008$ & $0.3117$ & \textbf{0.2494} & $0.6979$ & $0.1134$ & $0.1129$\\  & ViTs16 & $0.649$ & $0.6558$ & $0.2922$ & \textbf{0.2379} & $0.6502$ & $0.09539$ & $0.0904$\\  & ViTb8 & $0.7101$  & $0.7136$ & $0.3325$ & \textbf{0.2518} & $0.7104$  & $0.09856$ & $0.08615$\\ & ViTb16 & $0.6832$ & $0.6851$ & $0.3067$ & \textbf{0.2477} & $0.681$ & $0.1102$ & $0.1072$\\\hline

\hline
\end{tabular}
 \caption{Accuracy of our method compared to the Random\_debugging. Note that the accuracy of models on debug\_seed and debug\_heldout was zero before debugging. After applying our debugging method, we gain above $\sim 21 \%$ improvements in accuracies for all models, showcasing that more than $\sim 21 \%$ of model errors in the heldout set come from wrong background associations.}
\label{table:individual model acc}
\end{table*}

\subsubsection{Collective Failure Mitigation}
\label{section:debugging_all}
As discussed in section \ref{section:failure_all}, since models from the same categories have very similar failures, we have considered the possibility of using a single set of generated data, called \textbf{\textit{collective\_debug\_train}}, to debug all models within the same categories. To achieve this, we have devised two different settings: 1) we get the failure modes of all models in the same category (e.g. ResNets), and we select \textit{k} samples from all their failures. Therefore, background failures that occurred more have a higher probability of being chosen for the collective\_debug\_train. We then use this data to debug individual models in this category. 2) We get the failure modes of only one of the models in a category and then use this to debug all models. This approach is more efficient in terms of time and memory, as it requires running only one model per category.  The results for this experiment are shown in table \ref{table:all models acc}.
Based on our observations in section \ref{section:failure_all}, having the same debug\_train data for debugging (collective\_debug\_train), improves the accuracies among all models in the same category. This approach offers greater efficiency as it eliminates the need to generate debug\_train data for each individual model. Consequently, it saves us both time and memory that would otherwise be required for storing such data.

\begin{figure}
        \centering
        \includegraphics[trim=0cm 0cm 0cm 0cm, clip, width=0.6\linewidth]{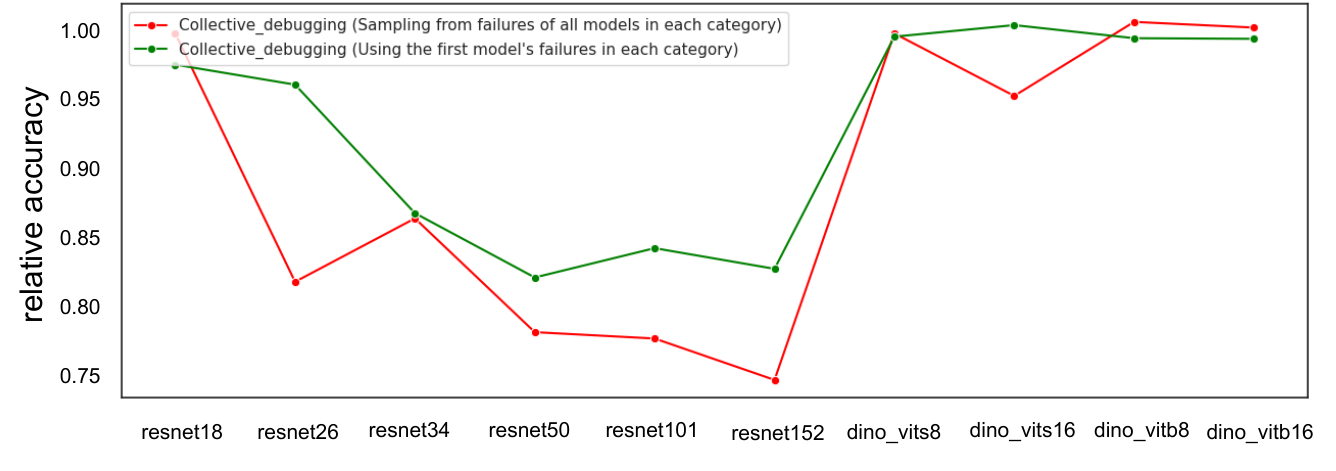}
    \caption{Comparison of resolved failures between collective\_debugging and individual\_debugging as a percentage. Relative accuracy is the ratio of the combined debugging's accuracy over individual debugging's accuracy on each specific model.} 
    \label{fig:90_plot}
\end{figure}

In an overview, the \textbf{collective} debugging approach showcases the capability to resolve above $\textbf{75\%}$ of failures corrected by individual debugging (Debugging each model based on its failures) \ref{fig:90_plot}.

\begin{table*}
    \centering
\begin{tabular}{||c|c|c|c|c|c|c|c|c|c||}
\hline
 Models & \multicolumn{8}{ |c|| }{Accuracies}\\ \hline
\hline
 \multirow{3}{*} {} & & Accuracy &\multicolumn{3}{ |c| }{\textbf{Accuracy of}} &\multicolumn{3}{ |c|| }{\textbf{Accuracy of}}\\
    Model & model & before & \multicolumn{3}{ |c| }{\textbf{Collective Debugging-type 1}} & \multicolumn{3}{ |c|| }{\textbf{Collective Debugging-type 2}}\\
     category & name & debugging & \multicolumn{3}{ |c| }{(ours)} & \multicolumn{3}{ |c|| }{(ours)}\\ \hline
  
 & & Test & Test & seed & \textbf{heldout} & Test & seed & \textbf{heldout}\\ \hline
\multirow{6}{*}{ResNet} & resnet18 & $0.6236$ & $0.6364$ & $0.2291$ & \textbf{0.2078} & $0.6413$ & $0.2636$ & \textbf{0.2128}\\  & resnet26 & $0.6593$ & $0.6655$ & $0.2396$ & \textbf{0.2192} & $0.6669$ & $0.2185$ & \textbf{0.1853}\\  & resnet34 & $0.7017$  & $0.7149$ & $0.2312$ & \textbf{0.2188} & $0.7135$ & $0.2391$ & \textbf{0.2178}\\ & resnet50 & $0.7631$ & $0.7644$ & $0.2419$ & \textbf{0.2217} & $0.7641$ & $0.2325$ & \textbf{0.2105}\\ & resnet101 & $0.796$ & $0.8001$ & $0.2474$ & \textbf{0.2226} & $0.7999$ & $0.2244$ & \textbf{0.2045}\\ & resnet152 & $0.816$ & $0.8182$ & $0.2509$ & \textbf{0.2317} & $0.8188$ & $0.2253$ & \textbf{0.2081}\\ \hline

 \multirow{6}{*}{DINO} & ViTs8 & $0.6977$ & $0.70001$ & $0.2729$ & \textbf{0.2488} & $0.70008$ & $0.3117$ & \textbf{0.2494}\\  & ViTs16 & $0.649$ & $0.6522$ & $0.2673$ & \textbf{0.2394} & $0.6504$ & $0.2563$ & \textbf{0.2267}\\  & ViTb8 & $0.7101$  & $0.7125$ & $0.2913$ & \textbf{0.2509} & $0.7117$ & $0.2903$ & \textbf{0.2540}\\ & ViTb16 & $0.6832$ & $0.6840$ & $0.2985$ & \textbf{0.2467} & $0.6833$ & $0.2855$ & \textbf{0.2488}\\ \hline
 
\hline
\end{tabular}
 \caption{Collective\_debug\_train method's results.}
\label{table:all models acc}
\end{table*}

\section{Conclusion}
In our project, we've developed a technique to identify failure modes by focusing on a specific category of spurious correlations. We then leverage these detected failures to generate additional samples, allowing the model to learn from and address its shortcomings. We've illustrated the resemblance of failures within a particular model category, highlighting that models with the same architecture share more similar failures. Exploiting this insight, we've devised a method to alleviate failures across all models in a category using a single set of generated data based on the failures of just one model in that category. Our results indicate that this collective debugging approach can resolve over $75\%$ of failures addressed through individual debugging efforts. Our framework empowers users to select the spurious correlation to identify and mitigate, facilitating the simultaneous debugging of a subset of models with a single (small) auxiliary set of additional data, thereby saving both time and resources.

\bibliography{main}
\bibliographystyle{tmlr}

\appendix
\newpage
\section{Appendix}

\begin{figure}[h]
        \centering
        \includegraphics[trim=0cm 0cm 0cm 0cm, clip, width=0.7\linewidth]{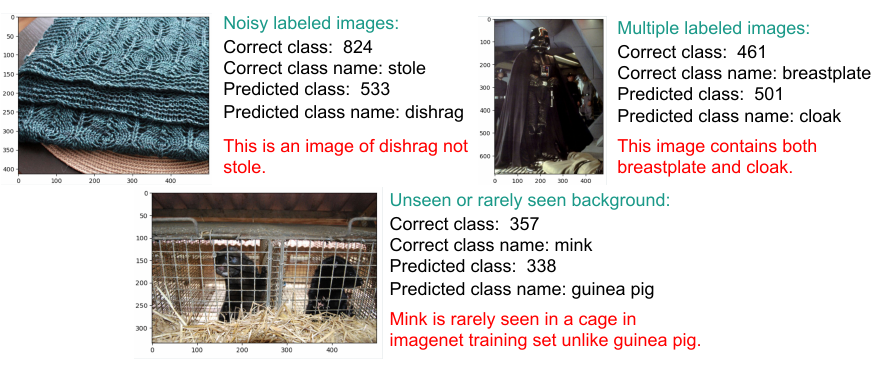}
    \caption{examples of 3 most common failure modes of deep learning models} 
    \label{appendix:failure_examples}
\end{figure}

\begin{table}[h]
    \centering
\begin{tabular}{||c|c||} 
 \hline
 Parameter & Value\\
 \hline \hline
 Leearning rate & 0.2\\ 
 \hline
 Epochs & 1000\\ 
 \hline
 Momentum & 0.9\\ 
 \hline
 Weight decay & 0.0005\\ 
 \hline
 \# Chosen common BGs & 5\\
 \hline
 Lambda & 0.5\\
 \hline
\end{tabular}
 \caption{Shared parameters among all dataset.}
\label{table:shared_params}
\end{table}

\begin{table*}
    \centering
\begin{tabular}{||c|c|c|c||}
\hline
\multicolumn{4}{ ||c|| }{Models} \\
\hline
 Model\_category & \multicolumn{3}{ |c|| }{Model\_name}\\ \hline
\multirow{2}{*}{ResNet} & resnet18 & resnet26 & resnet34\\
& resnet50 & resnet101 & resnet152 \\ \hline

\multirow{4}{*}{EfficientNet} & efficientnet\_b0 & efficientnet\_b1 & efficientnet\_b2 \\
& efficientnet\_b3 & efficientnet\_b4 & efficientnet\_b5 \\
& efficientnet\_b6 & efficientnet\_b7
& efficientnet\_b8 \\
& efficientnet\_l2 & & \\   \hline

\multirow{1}{*}{DenseNet} & densenet121 & densenet161 & \\ \hline

\multirow{3}{*}{ViT} & vit\_base\_patch16\_224 & vit\_base\_patch32\_224 & vit\_large\_patch16\_224 \\
& vit\_large\_patch32\_224 & vit\_base\_resnet26d\_224 & vit\_base\_resnet50d\_224 \\ \hline

\multirow{2}{*}{SWaV} & resnet50 & resnet50w2 & resnet50w4 \\
& resnet50w5 & & \\ \hline

\multirow{1}{*}{MoCo} & moco\_v2\_800ep &  &  \\ \hline

\multirow{2}{*}{DINO} & dino\_resnet50 & dino\_vitb16 & dino\_vitb8 \\
& dino\_vits16 & dino\_vits8 & \\ \hline

\multirow{2}{*}{CLIP} & ViT-B\/32 & RN50 & RN101 \\
& ViT-L\/14 & RN50x4 & RN50x16\\ \hline
 
\hline
\end{tabular}
 \caption{List of models we tested our debugging framework on.}
\label{table:models_list}
\end{table*}

\newpage
\section{Results for all models on ImageNet}
\label{appendix:B}

\begin{table}
    \centering
\begin{tabular}{||c|c|c|c|c|c|c|c|c||}
\hline
 Models & \multicolumn{8}{ |c|| }{Accuracies}\\ \hline
\hline
 \multirow{3}{*} {Model} & & Accuracy & \multicolumn{3}{ |c|| }{\textbf{Accuracy of}} & \multicolumn{3}{ |c|| }{Accuracy of}\\
    & model & before  & \multicolumn{3}{ |c|| }{\textbf{Individual Debugging}} & \multicolumn{3}{ |c|| }{Random debugging}\\
  category & name & debugging  & \multicolumn{3}{ |c|| }{(ours)} & \multicolumn{3}{ |c|| }{}\\\hline

 & & Test & Test & seed & \textbf{heldout} & Test & seed & heldout \\ \hline
\multirow{10}{*}{EfficientNet} & b0 & $0.715$ & $0.7198$ & $0.2034$ & \textbf{0.1842} & $0.7134$ & $0.0894$ & $0.0883$\\  & b1 & $0.7373$ & $0.74415$ & $0.2154$ & \textbf{0.1885} & $0.7399$ & $0.1003$ & $0.1011$\\  & b2 & $0.7525$  & $0.7591$ & $0.2283$ & \textbf{0.1909} & $0.7448$  & $0.0957$ & $0.0942$\\ & b3 & $0.7634$ & $0.7730$ & $0.2318$ & \textbf{0.1991} & $0.7669$ & $0.1066$ & $0.1010$\\ & b4 & $0.7701$ & $0.7780$ & $0.2398$ & \textbf{0.2068} & $0.7719$ & $0.0955$ & $0.0960$\\ & b5 & $0.7821$ & $0.7819$ & $0.2405$ & \textbf{0.2055} & $0.7761$ & $0.0893$ & $0.0915$\\ & b6 & $0.7884$ & $0.7886$ & $0.2561$ & \textbf{0.2083} & $0.7863$ & $0.0941$ & $0.0914$\\ & b7 & $0.7895$ & $0.7903$ & $0.2600$ & \textbf{0.2126} & $0.7898$ & $0.0951$ & $0.0972$\\  & b8 & $0.7928$ & $0.7951$ & $0.2653$ & \textbf{0.2147} & $0.7932$ & $0.0972$ & $0.0934$\\ \hline

 \multirow{2}{*}{DenseNet} & 121 & $0.6792$ & $0.6869$ & $0.2138$ & \textbf{0.1592} & $0.6773$ & $0.0651$ & $0.0664$\\  & 161 & $0.7254$ & $0.7332$ & $0.2418$ & \textbf{0.1833} & $0.7249$ & $0.0779$ & $0.0771$\\ \hline

 \multirow{6}{*}{ViT} & base\_patch16\_224 & $0.739$ & $0.7477$ & $0.2501$ & \textbf{0.2193} & $0.7399$ & $0.1047$ & $0.1044$\\  & base\_patch32\_224 & $0.7456$ & $0.7493$ & $0.2574$ & \textbf{0.2199} & $0.7469$ & $0.1078$ & $0.1072$\\  & large\_patch16\_224 & $0.7493$  & $0.7539$ & $0.2644$ & \textbf{0.2263} & $0.7468$  & $0.0952$ & $0.0957$\\ & large\_patch32\_224 & $0.7535$ & $0.7545$ & $0.2674$ & \textbf{0.2274} & $0.7553$ & $0.1023$ & $0.1041$\\ \hline

 \multirow{4}{*}{SWaV} & resnet50 & $0.4254$ & $0.4384$ & $0.1403$ & \textbf{0.1274} & $0.4267$ & $0.0662$ & $0.0624$\\  & resnet50w2 & $0.4317$ & $0.4328$ & $0.1583$ & \textbf{0.1294} & $0.4319$ & $0.0683$ & $0.0652$\\  & resnet50w4 & $0.4402$  & $0.4477$ & $0.1592$ & \textbf{0.1304} & $0.4416$  & $0.0672$ & $0.0617$\\ & resnet50w5 & $0.4526$ & $0.4589$ & $0.1633$ & \textbf{0.1363} & $0.4552$ & $0.0696$ & $0.0703$\\ \hline

 \multirow{1}{*}{MoCo} & v2\_800ep & $0.6931$ & $0.6946$ & $0.2041$ & \textbf{0.1584} & $0.6937$ & $0.0943$ & $0.0917$\\ \hline

\hline
\end{tabular}
 \caption{Accuracy of our method comparing to the debug random. Note that the accuracy of models on debug\_seed and debug\_heldout was zero before debugging. These results are for all models that we tested our method on.}
\label{table:appendix_B}
\end{table}

\end{document}